%% file: main.tex
\documentclass[10pt,conference]{IEEEtran}

\input{macros.tex}

\usepackage{hyperref}
\usepackage{graphicx}
\usepackage{amsmath}
\usepackage{amssymb}
\usepackage{tabularx}
\usepackage{graphics}
\usepackage{url}
\usepackage{paralist}
\usepackage{orcidlink}
\usepackage{adjustbox}

\begin{document}
\title{An Adaptive Clustering Scheme for Client Selections in Communication-efficient Federated Learning}

\author{
    \IEEEauthorblockN{
        Yan-Ann~Chen\orcidlink{0000-0002-3348-6022} and Guan-Lin~Chen
    }
    \IEEEauthorblockA{
    Dept. of Computer Science and Engineering, Yuan Ze University, Taoyuan, Taiwan
    }
}

\maketitle

\begin{abstract}
Federated learning is a novel decentralized learning architecture. During the training process, the client and server must continuously upload and receive model parameters, which consumes a lot of network transmission resources. Some methods use clustering to find more representative customers, select only a part of them for training, and at the same time ensure the accuracy of training. However, in federated learning, it is not trivial to know what the number of clusters can bring the best training result. Therefore, we propose to dynamically adjust the number of clusters to find the most ideal grouping results. It may reduce the number of users participating in the training to achieve the effect of reducing communication costs without affecting the model performance. We verify its experimental results on the non-IID handwritten digit recognition dataset and reduce the cost of communication and transmission by almost 50\% compared with traditional federated learning without affecting the accuracy of the model.
\end{abstract}


\section{Introduction} \label{Sec:Intro}
\emph{Federated learning (FL)} \cite{mcmahan2017communication} draws a great attention to applications that need privacy-aware training such as Google Keyboard \cite{yang2018applied}, medical image recognition \cite{antunes2022federated}, and Internet of Things \cite{zhang2021federated}.
This learning paradigm enables multiple participants (also called clients) to collaboratively train a machine learning model by only exchanging model parameters with a server.
The server collects parameters from all clients, aggregates them to create a new global model, and returns them to all clients.
This approach ensures the privacy of client data since clients do not have to upload the original data to the server.
In addition, FL benefits from the improved performance achieved by all the knowledge of the clients.

When exploiting the FL framework, we face certain challenges with respect to the properties of FL.
In traditional machine learning, we collect all data toward a centralized server, and we can also perform a preprocessing step to balance the training data for multiclass classification.
But in FL, each client trains their model collaboratively without revealing their data and its class distribution.
Clients of an FL application may have biases to the global population because of their preferences, deployment places, or tasks, where we call this situation non-IID (non-identically independently distributed).
Moreover, these clients have to distributively train the model via multiple iterations which incur a large amount of data exchange.
Several research works focus on the use of the clustering method \cite{ouyang2021clusterfl,FedSAUC,chen2021similarity} to select participating clients to reduce data exchange while maintaining the accuracy of the model in the non-IID setting.
However, reference \cite{FedSAUC} has the strong assumption that the number of clusters for all clients is known in advance.

In this work, we investigate the use of adaptive clustering to select clients to reduce communication costs and let the global model contain adequate representative data.
Data from users with similar behavior or usage may provide similar learned features.
In the system, all clients can be divided into clusters according to their similarity of data (features).
To the best communication efficiency, we may just let one of them in each cluster join the FL training.
However, we cannot estimate the possible number of clusters because the server does not have all the data inside each client, and the client also does not know the condition of others.
Therefore, we propose adaptive clustering methods and two stabilization techniques to estimate the number of clusters.

\section{System Architecture and Design} \label{Sec:Methods}

\begin{figure}[!t]
\centering
\includegraphics[width=0.95\linewidth]{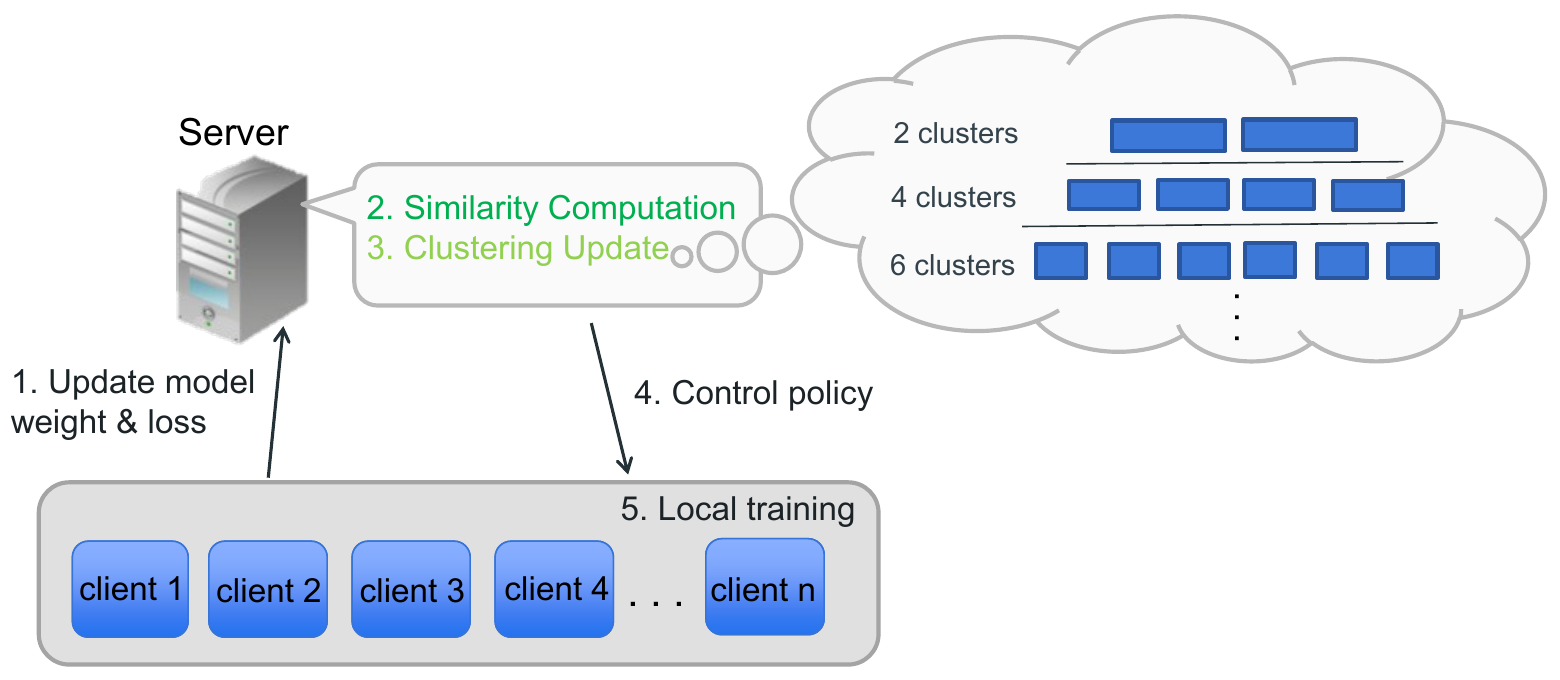}
\caption{System Architecture.}
\label{Fig:FL_arch}
\end{figure}
\textbf{System Architecture.}
\Fig{Fig:FL_arch} shows the system architecture. The system flow follows the well-known federated learning framework, FedAvg \cite{mcmahan2017communication}, and we add the components of similarity computation and cluster-based client selections.
Before applying the client selection, we need all the clients send their models and losses onto the parameter server. The server will update and send back the global model.
After several rounds of training, we start to compute the similarity between each pair of clients according to their local models.
Then we divide these clients into the number of clusters which computes by our proposed method.
For each cluster, we may select some representatives to join the following training.
We will apply this clustering process for each round and update the participant list to all the clients.

\textbf{Adaptive Clustering.}
In this work, we adopt agglomerative hierarchical clustering \cite{murtagh2012algorithms} since we can easily compute different clustering results by adjusting the distance threshold.
Thus, after we compute the model similarity of each pair of clients, we can iteratively merge the clusters of the closest two clients into one cluster until there is only one rest of the cluster.
Then we may determine the distance threshold such that the number of clusters is equal to our decided one.

We refer to the concept of TCP flow control and propose a dynamic adjustment method to determine the number of clusters. Initially, we set the number of clusters $p$ to the total number of clients $n$. The server collects client models and their losses iteratively. For each round $i$ ($i = 1..m$), it computes the average losses $L_i$ from all clients. If the reduction ratio between $L_{i-1}$ and $L_i$ is greater than a threshold $w$, we decrease the number of clusters $p$ by $d$. At first, $d$ is equal to $1$ and $d$ increases exponentially each round until $p = 1$ if the condition is satisfied during these consecutive rounds. Otherwise, $p$ goes back to $2*p$ ($n$ is the maximum number). To stabilize $p$, we may let $p$ remain unchanged for a number of rounds.

Although we may keep it stable for a few rounds, the TCP-like method will change the number of clusters after the stabilizing period. Therefore, we propose two other stabilizing mechanisms such that the number of clusters stays the same for a longer period.
\begin{itemize}
  \item SA-like: We refer to the simulated annealing algorithm for the optimization problem. We set a probability to let the number of clusters remain unchanged when the loss reduction ratio is below the threshold. On the other hand, we still reduce the number of clusters if we have a satisfied reduction ratio.
  \item Experience-based: As SA-like method, we have a probability of keeping the number of clusters. Here, the probability is determined by the good/bad experiences when previously the clustering setting stayed in this number.
\end{itemize}

\section{Experimental Results}
In our experiment, we use FedML \cite{fedml} to implement FL training and use MNIST \cite{mnist}, which is a handwritten digit recognition dataset, to carry out the training.
To establish a non-IID situation, we intentionally partitioned the data for each client.
In the training, we had a total of $8$ clients participating in federated learning.
Each pair of clients received a dataset with two specific digit labels.
For example, clients 0 and 1 have data with labels 0 and 1, while clients 2 and 3 have data with labels 2 and 3, and so on.
In such an extremely non-IID scenario, we anticipated that the ideal number of clusters would be $4$.

For the machine learning model, we adopt a simple convolutional neural network (CNN) model. The CNN structure consists of $2$ convolutional layers, each followed by a ReLU activation function and max pooling.
The output of these layers is then flattened and connected to $2$ fully connected layers.
We provide detailed specifications for the internal layers in the following. The first and second convolutional layers have a kernel size of $3$x$3$, with $32$ and $64$ kernels, respectively. Both max pooling operations have a size of $2$x$2$. After flattening, the first fully connected layer contains $128$ neurons and the second fully connected layer has $8$ units for classification.

For performance comparison, we evaluate models by accuracy and network communication costs.
We assume that each model transmission carries the same number of parameters and ignore the time and distance for transmitting and receiving the model between each client and the server. Thus, we only compare the differences in the accumulated transmission count during training.

\begin{table}[t!]
\centering
\caption{}
\label{tab:comparison}
\begin{adjustbox}{width=\linewidth,center}
\begin{tabular}{|c|c|c|}
  \hline
  Method & Top accuracy ($200$ rounds) & Transmission times\\
  \hline
  \hline
  FedAvg & $96.16\%$ & $1600$ \\
  FedSAUC(4) & $96.19\%$ & $808$ \\
  FedSAUC(2) & $95.38\%$ & $808$ \\
  FedSAUC(1) & $95.21\%$ & $808$ \\
  Our with SA-like & $95.95\%$ & $801$ \\
  Our with experiences & $96.2\%$ & $802$ \\
  \hline
\end{tabular}
\end{adjustbox}
\end{table}

\Tab{tab:comparison} compares FedAvg \cite{mcmahan2017communication} and FedSAUC \cite{FedSAUC} with fixed numbers of clusters $1$, $2$, and $4$.
After the second round of training, FedSAUC fixes the number of clusters, randomly selecting half of the clients in each cluster.
On the other hand, our method randomly selects one client from each cluster for training.
We can observe that our method achieves the highest accuracy among the observed $200$ training rounds, which is closest to the ideal clustering number $4$.
For transmission times, our dynamic clustering method, which selects one client per cluster for training, even has a lower total transmission count compared to FedSAUC(4).
Therefore, our method can automatically search for the optimal clustering without prior knowledge of the ideal number of clusters in the non-IID setting of federated learning, reducing communication costs by nearly 50\% compared to the conventional federated learning framework FedAvg.

\section{Conclusions}\label{Sec:conc}
We have presented a dynamic clustering mechanism for communication-efficient federated learning. By considering the cost of communication, our approach dynamically adjusts the grouping of clients, selecting representative participants for each training round. Through hierarchical clustering and the use of stabilizing mechanisms, we may approach the number of clusters at the optimal value, reducing the need for manual exploration and preserving the diversity of the training samples. Our research demonstrates that by intelligently selecting representative clients and dynamically adjusting the clustering, our approach achieves a substantial reduction in communication cost in federated learning. The proposed mechanism shows its potential to handle non-IID data settings without prior knowledge of the ideal clustering number. In general, our work contributes to the advancement of efficient and scalable federated learning techniques.

\section*{Acknowledgement}
\thanks{
    Y.-A. Chen's research
    is co-sponsored by
    MOST 107-2218-E-155-007-MY3, Taiwan, 
    MOST 110-2221-E-155-022-MY3, Taiwan, 
    and Innovation Center for Big Data and Digital Convergence, Yuan Ze University, Taiwan.
}

\bibliographystyle{IEEEtran}
\bibliography{bibfile}
\end{document}

%% file: macros.tex
\newcommand{\blist}{\begin{list}\setlength{\topsep 0pt \parsep 0pt}}
\newcommand{\elist}{\end{list}}
\newtheorem{thm}{Theorem}
\newcommand{\bthm}{\begin{thm}\begin{textit}}
\newcommand{\ethm}{\end{textit}\end{thm}}
\newtheorem{dfn}{Definition}
\newcommand{\bdfn}{\begin{dfn}\begin{textit}}
\newcommand{\edfn}{\end{textit}\end{dfn}}
\newtheorem{obn}{Observation}
\newcommand{\bobn}{\begin{obn}\begin{textit}}
\newcommand{\eobn}{\end{textit}\end{obn}}

\newtheorem{lema}{Lemma}
\newcommand{\bdla}{\begin{lema}\begin{textit}}
\newcommand{\edla}{\end{textit}\end{lema}}
\newtheorem{pro}{Property}
\newcommand{\bpro}{\begin{pro}\begin{textit}}
\newcommand{\epro}{\end{textit}\end{pro}}

\newcounter{algnum1}

\newcounter{algnum2}

\newcounter{algnum3}

\newcommand\bit{\begin{itemize}}
\newcommand\eit{\end{itemize}}

\newcommand{\Fig}[1]{Fig.~\ref{#1}}

\newcommand{\Tab}[1]{Table~\ref{#1}}

\newcommand{\balgo}{
   \begin{description}
   \parskip=0pt
   \topsep=0pt
}
\newcommand{\ealgo}{
   \end{description}
}